\documentclass[10pt,twocolumn,letterpaper]{article}

\usepackage{cvpr}              
\definecolor{cvprblue}{rgb}{0.21,0.49,0.74}
\usepackage[pagebackref,breaklinks,colorlinks,allcolors=cvprblue]{hyperref}
\usepackage{algorithm}
\usepackage{algorithmic}
\usepackage{amsfonts}
\usepackage{amsmath}
\usepackage[table,xcdraw]{xcolor}
\usepackage{threeparttable}
\usepackage{booktabs}
\usepackage[accsupp]{axessibility}  

\def\paperID{xxxx} 
\def\confName{CVPR}
\def\confYear{2026}

\title{Open-Ended Instruction Realization with LLM-Enabled Multi-Planner Scheduling in Autonomous Vehicles}

\author{Jiawei Liu$^{1,6}$\quad Xun Gong$^{1,6}$\thanks{Corresponding author}\quad Fen Fang$^{2}$\quad Muli Yang$^{2}$\quad Bohao Qu$^{2}$\quad Yunfeng Hu$^{1}$\quad Hong Chen$^{3}$\\
Xulei Yang$^{2}$\quad Qing Guo$^{4,5}$
\\
$^{1}$Jilin University, China\quad
$^{2}$Agency for Science, Technology and Research (A*STAR), Singapore\\
$^{3}$ Tongji University, China\quad
$^{4}$ NKIARI, China\quad
$^{5}$ Nankai University, China\\
$^{6}$ Engineering Research Center of Knowledge-Driven Human-Machine Intelligence, MOE, China
}

\begin{document}
\maketitle
\begin{abstract}
Most Human-Machine Interaction (HMI) research overlooks the maneuvering needs of passengers in autonomous driving (AD). Natural language offers an intuitive interface, yet translating passenger open-ended instructions into control signals—without sacrificing interpretability and traceability—remains a challenge. This study proposes an instruction-realization framework that leverages a large language model (LLM) to interpret instructions, generates executable scripts that schedule multiple model predictive control (MPC)-based motion planners based on real-time feedback, and converts planned trajectories into control signals. This scheduling-centric design decouples semantic reasoning from vehicle control at different timescales, establishing a transparent, traceable decision-making chain from high-level instructions to low-level actions. Due to the absence of high-fidelity evaluation tools, this study introduces a benchmark for open-ended instruction realization in a closed-loop setting. Comprehensive experiments reveal that the framework significantly improves task-completion rates over instruction-realization baselines, reduces LLM query costs, achieves safety and compliance on par with specialized AD approaches, and exhibits considerable tolerance to LLM inference latency. For more qualitative illustrations and a clearer understanding.
\end{abstract}    
\section{Introduction}
\label{sec:intro}

\begin{figure*}[!t]
\centering
\includegraphics[width=2.0\columnwidth]{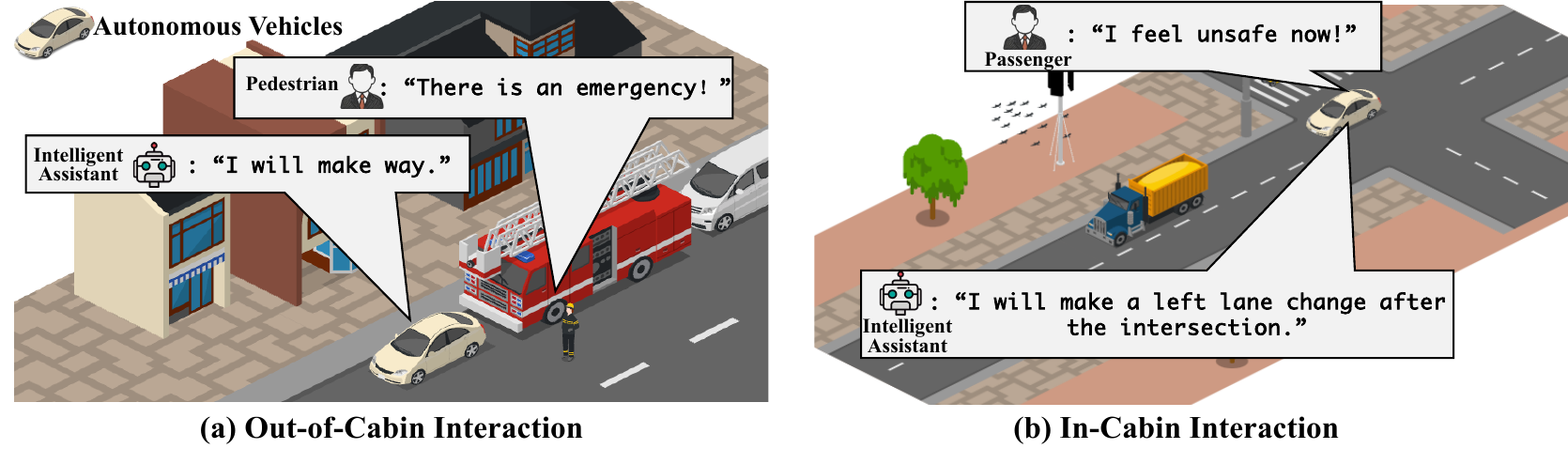} 
\caption{Human-Like HMI Using Open-Ended Language Instructions. Here, “open-ended” refers to diverse natural-language phrasings.}
\label{fig:HMI}
\end{figure*}

Existing Human-Machine Interaction (HMI) systems, developed primarily for Society of Automotive Engineers (SAE) Levels 0-3, are insufficient for L4-L5 autonomous driving (AD) systems such as Robotaxi. Prevailing HMIs (e.g., lane-departure warning at L0 \cite{LaneDepartureWarning}, lane-keeping assistance at L1-L2 \cite{LaneKeepingAssistant}, and takeover control at L3 \cite{TakeOverControl}) presume a driver ready to intervene. As shown in Figure \ref{fig:HMI}, AD research shifts attention to broader traffic participants such as rear-seat passengers \cite{HMISurvey}. Driver-oriented cues, like flashing lights, haptic steering feedback, and takeover alerts, therefore lose relevance. This transition requires redesigning HMI for intuitive interaction with non-driving users.\par

Recent advances in Large Language Model (LLM) offer a compelling path forward \cite{LLMAVSurvey}. Achieving a human-like interaction has long been a goal of onboard HMI \cite{HMISurvey}. Natural language, the most universal medium of communication, offers clear advantages. Meanwhile, LLMs excel at comprehending language input and producing understandable responses, making them well-suited for enabling bidirectional interaction between passengers and vehicles.\par

LLM-driven HMI has gained momentum in research and industry in the last two years. Recent AD methods leveraging LLM agents and VLA models exemplify language–driving integration \cite{LLMAVSurvey,VLA4ADSruvey}. Industrial progress mirrors this trend. In May 2025, Apple introduced CarPlay Ultra, allowing drivers to control cabin climate and audio through Siri \cite{AppleNews}. That same month, Li Auto unveiled its “Driver Agent” concept, identifying it as a focus for future research \cite{LixiangNews}. The global investment bank, Goldman Sachs, projects China’s Robotaxi market to increase from USD 54 million in 2025 to USD 12 billion by 2030 \cite{GoldmanSachs}.\par

Despite this momentum, key challenges continue to hinder making language the primary mode of human–vehicle interaction and supplanting the century-old steering wheel, accelerator, and brake. \textit{\textbf{Challenge A: Underexplored Open-Ended and Maneuver-Level Instruction}.} Real-world passenger instructions vary culturally, and rarely follow standard templates. Meanwhile, current onboard HMIs emphasize infotainment, cabin control, and route guidance (or navigation), but offer limited access to driving maneuvers such as lane change, overtaking, or pulling over \cite{HMISurvey,Tesla}. Interpreting and executing open-ended, maneuver-level instructions (Figure \ref{fig:HMI}) remains an underexplored problem in designing human-centric interaction systems. \textit{\textbf{Challenge B: Lack of Efficient Behavior Scheduling}.} Understanding intent is essential, but executing instructions adds more complexity, requiring the scheduling of multiple driving behaviors. For example, the instruction “\textit{I feel unsafe}” in Figure \ref{fig:HMI} invokes a behavior sequence—[\textit{left lane change}, \textit{acceleration}, \textit{lane keeping}]—each with distinct goals that cannot be managed by a single planner. To accurately perform instruction tasks in evolving traffic, behavior scheduling or switching must operate concurrently based on real-time feedback, without blocking other AD modules. \textit{\textbf{Challenge C: Insufficient High-Fidelity and Closed-Loop Evaluation}.} Most LLM-based AD research relies on open-loop evaluation using public datasets (e.g., Argoverse \cite{Argoverse}, NuScenes\cite{Nuscenes}) or game-style simulators (e.g., Highway\cite{Highway}, Carla\cite{Carla}), while closed-loop evaluation in hybrid simulations built on realistic traffic data remains uncommon \cite{LLMAVSurvey}. 


To tackle these issues, this study introduces an LLM-enabled, scheduling-centric framework. First, the LLM interprets open-ended instructions, resolves ambiguities by referencing traffic context, and outputs a driving behavior sequence. Next, it produces a script that schedules multiple motion planners to carry out the behavior sequence, integrating coroutine mechanisms \cite{coroutines} and asynchronous triggers to enable adaptive planner switching within evolving traffic. Last, MPC-based motion planner and dedicated controllers are employed to generate continuous control signals. This scheduling-centric architecture confines language-trained model's involvement to \textit{high‑level, low‑frequency} semantic reasoning, while a real-time feedback‑driven schedule–plan–control loop enforces \textit{low‑level, high‑frequency} safe adaptation, establishing a transparent and traceable decision-making chain from language instructions to numerical control signals. The main contributions are summarized as follows: 
\begin{itemize}
    \item \textbf{POINT Benchmark}: Due to the lack of testbeds for open-ended instructions, POINT augments the hybrid nuPlan simulator \cite{Nuplan} with 1,050 instruction–scenario pairs, enabling high-fidelity, closed-loop evaluation in simulated urban traffic. It also categorizes current LLM-based AD methods through a task-scheduling perspective and introduces several competitive baselines.

    \item \textbf{Scheduling-Centric Framework}: The proposed framework leverages the LLM's scheduling capability to coordinate explicit motion planners, enabling open-ended, maneuver-level instruction realization while maintaining a transparent language-to-control chain.

    \item \textbf{Comprehensive Evaluation}: This work compares the proposed framework with LLM-based, data-driven, and rule-based methods across various metrics. It outperforms instruction-realization baselines by 64\%-200\% with a single LLM query, matches safety and compliance standards of leading specialized AD methods, and exhibits considerable tolerance to LLM inference latency.
\end{itemize}

\section{Related Work}

This section briefly reviews instruction-processing approaches and VLA methods.\par

\subsection{Conventional Methods}
Conventional instruction processing typically adopts a two-stage pipeline: intent classification followed by key parameter extraction (e.g., speed, cabin temperature, destination) \cite{CommandSurvey, HMISurvey}. Approaches are typically either \textit{rule-based} or \textit{data-driven}. Rule-based systems handcraft grammars and templates to capture frequent instruction patterns, whereas data-driven systems learn classifiers for intent recognition. For instance, SpatialRoutines \cite{SpatialRoutines} uses a manually designed grammar to parse commands into spatial-routine scripts that guide a robot through a simulated maze. AIME \cite{AIME} collects multi-turn human-vehicle dialogs and trains separate RNNs for intent classification and key parameter extraction. A hierarchical framework \cite{OldConditionalLanguageControl} introduces a gated-attention encoder to convert commands into conditional inputs for a policy network, enabling language-guided control.

While effective for limited, standardized commands, these approaches are brittle in the Robotaxi setting: (i) \emph{Open-Domain Mismatch}: Applying rule-based methods to passenger instructions requires enumerating massive rules over the cross-product of open language, driving scenes, and continuous vehicle actions (see Section~\ref{sec: formulation}), leading to combinatorial explosion, sparse coverage, and high maintenance \cite{IntentClassification}. (ii) \emph{Intent-Slot Rigidity}: Data-driven intent classification relies on a fixed set of labels and phrasing in training data, limiting OOD generalization \cite{SlotFilling}. Moreover, key parameter extraction assumes clearly specified parameters (or slots), which passenger instructions often violate.\par

In this work, passenger instructions are only used for evaluation, with neither template/rule derivation nor model training to avoid data leakage. Thus, the benchmark aims to evaluate open-ended instruction realization rather than assess rule coverage or in-distribution generalization.\par

\begin{figure*}[!t]
\centering
\includegraphics[width=2.0\columnwidth]{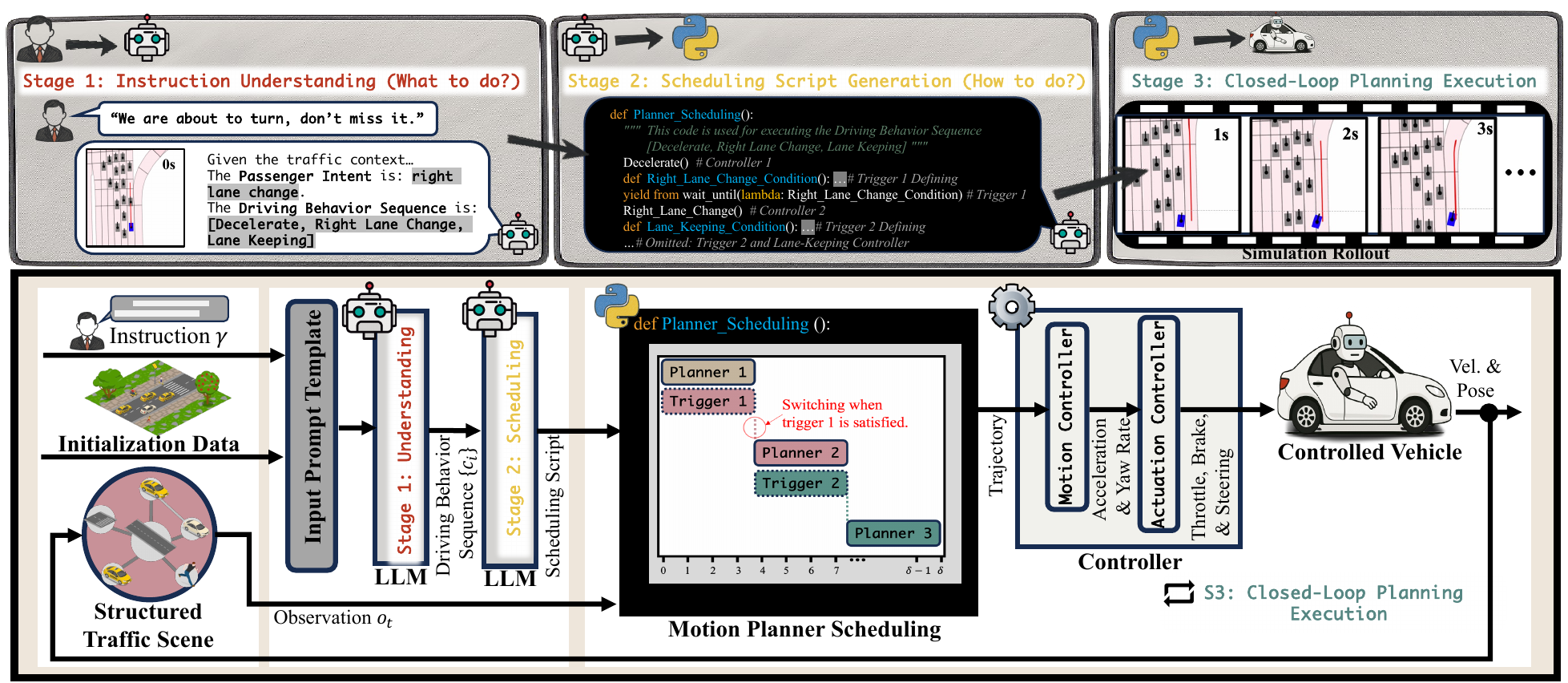} 
\caption{A Scheduling-Centric Framework Powered by LLM. It leverages an LLM to interpret passenger instruction and generate an executable script. The script then schedules multiple motion planners to realize the instruction using real-time feedback.}
\label{fig:framework}
\end{figure*}

\subsection{LLM-based Solutions}

Recent vision–language–action (VLA) methods augment vision–language models (VLMs) with action heads or experts, unifying perception, language, and control to exploit internet-scale knowledge and powerful reasoning for enhanced driving performance \cite{VLA4ADSruvey}. For instance, LMDrive \cite{LMdrive} consumes camera frames and navigation commands to control steering, throttle, and brake. AutoVLA \cite{AutoVLA} discretizes trajectories into physically feasible tokens to generate executable plans from multi-view image and language. AdaThinkDrive \cite{AdaThinkDrive} adaptively determines, based on scene complexity, whether to perform reasoning before planning.\par

Nonetheless, prior work faces the following gaps when applied to open-ended instructions: (i) \emph{Open-ended Instruction Understanding}: Most recent work focus on driving performance gain from language modality. This line of research therefore favors standardized, well-specified navigation commands like ``\textit{turn left at next intersection}" \cite{LMdrive}, or standardized queries like ``\textit{what is the next action?}" \cite{Alphadrive,DiLu} and ``\textit{predict the future trajectory in the next three seconds}" \cite{LM4TS_2, AdaThinkDrive}. This emphasis leads to the overlooking of open-ended instructions common in Robotaxi. Our study shows that interpreting such instructions is non-trivial: it requires both advanced reasoning of high-capacity LLMs and explicit use of traffic context to disambiguate intent. (ii) \emph{Language–Action Traceability}: While end-to-end design curbs error accumulation and information loss, it can reduce VLA transparency \cite{Traceability}. Recent studies indicate that textual reasoning (what VLAs say) and executed actions (what VLAs do) are not always closely aligned \cite{ReasonActionMisalignment,Libero,Alpamayo-R1}. This inconsistency further complicates the compliance with safety standards such as ISO 26262, which encourage a traceable and transparent decision-making chain \cite{Traceability}.\par

To tackle the above gaps, this work proposes a scheduling-centric framework enlightened by control-theoretic design principles including hierarchical decoupling \cite{hierarchical_decoupling}, timescale separation \cite{timescale_separation}, and event-triggered scheduling \cite{event_triggered_scheduling}. This framework aims to exploit each component’s strength: the language-trained LLM interprets open-ended instruction and schedules explicit motion planners through high-level textual reasoning, while the optimization-driven, MPC-based planner manages low-level continuous-valued control.\par

\section{Problem Formulation}
\label{sec: formulation}
Passenger instruction realization can be formulated as an instruction-guided Partially Observable Markov Decision Process (POMDP), defined by 
\begin{equation}
\langle S, A, O, T, \mathcal{O}, \Gamma, R \rangle, 
\end{equation}
where $s_{t}\in S$ is the state at time $t$, $a_{t}\in A$ the action taken, $o_{t}\in O$ the observation received. The state transition function is $T(s'|s,a)$, and $\mathcal{O}(o|s)$ the observation model. As $s_t$ is not fully observable, the agent maintains a belief state $b_t:=\mathbb{P}(s_t\!=\!s \mid \xi_t)$, where $\xi_t:=\{o_0,a_0,\dots,a_{t-1},o_t\}$ is the observation-action history.\par

Given an instruction $\gamma\in\Gamma$, an interpreter $f_\phi$ should infer its intended task and map it into an ordered atomic driving behavior sequence (subtasks) $\left\{c_{i}\right\}^{m(\gamma)}_{i=1}$ where $m(\gamma)\in\mathbb{N}_{>0}$. Each behavior $c_{i}$ is associated with a completion set $\mathcal{C}_i\!\subseteq\!S$ and is considered complete when $s_t\!\in\!\mathcal{C}_i$.\par

Let $k_t\!\in\!\{0,\dots,m(\gamma)\}$ be the number of completed behaviors at time $t$ and define the augmented state $\bar s_t=(s_t,k_t)$. Sequential progress through the driving behavior sequence can be rewarded by:
\begin{equation}
R(\bar s_t,a_t,\bar s_{t+1})=
\begin{cases}
r_{k_t+1}, & k_{t+1}=k_t+1,\\
0, & \text{otherwise}
\end{cases}
\end{equation}
with
\begin{equation}
k_{t+1}=
\begin{cases}
k_t+1, & s_{t+1}\in\mathcal{C}_{k_t+1},\\
k_t, & \text{otherwise}.
\end{cases}
\end{equation}\par

To jointly optimize the interpreter $f_\phi$ and a policy $\pi_\theta$, the objective is to maximize the expected cumulative reward while ensuring safety:
\begin{equation}
\label{eq:obj}
\begin{aligned}
\max_{\phi, \theta} \ &  \mathbb{E}_{\gamma\sim \mathcal{P}(\Gamma)} \left[ \sum_{t=0}^{\delta(\gamma)} \mathbb{E}_{a_{t}\sim \pi_{\theta}(\cdot \mid b_t)} R(\bar{s}_{t}, a_{t}, \bar{s}_{t+1}) \right]  \\
\text{s.t.} \quad & \mathbb{P} \left[ \forall t\le\delta(\gamma):s_{t} \in S_{\text{safe}} \,\middle|\, b_0 \right] \geq 1 - \varepsilon,
\end{aligned}
\end{equation}
where $\delta(\gamma)$ is the task time horizon, $S_{\mathrm{safe}}$ is the set of admissible states, and $\varepsilon$ bounds the acceptable risk.

\section{Methodology}
\label{sec:method}

Equation (\ref{eq:obj}) is challenging due to discrete instruction parsing, sparse stage-wise rewards, and safety constraints. This study therefore introduces a scheduling-centric framework that resorts to LLM capabilities (see Figure \ref{fig:framework}).

\subsection{Instruction Intent Inference}
\label{sec:MethodIntent}
Given an instruction $\gamma$ and a textual scene description $o_{0}$, the framework takes LLM as the interpreter $f_{\phi}$ to infer the instruction intent and map it into a driving behavior sequence via $f_{\phi}(\gamma, o_{0})=\left\{c_{i}\right\}^{m(\gamma)}_{i=1}$. Each $c_{i}$ represents one of the five predefined atomic behaviors: \emph{lane keeping}, \emph{left lane change}, \emph{right lane change}, \emph{accelerate}, and \emph{brake}. The context $o_0$ resolves ambiguities by providing environmental constraints (e.g., disallowing a right lane change from the rightmost lane) and situational cues (e.g., initiating a lane change to pull over when necessary).\par

Interpreting open-ended instructions with LLM offers several advantages: (i) Semantic Reasoning: Pretrained world knowledge enables intent inference for OOD instructions via analogical \cite{Analogical} and compositional generalization \cite{Compositional}. (ii) Hallucination Mitigation: Textual scene description reduce the risk of hallucination compared to visual features. Additionally, constraining outputs to a structured sequence of predefined behaviors further anchors the response, enhancing reliability.\par

However, this reliance does not imply that the scheduling-centric framework operates on a purely text-represented traffic. Encoding fine-grained traffic cues (e.g., road geometry) as text inevitably loses detail, making textual scene descriptions ill-suited for safety-critical vehicle control in complex urban traffic. In the framework, the fast schedule–plan–control loop operates directly on raw perception inputs. The scheduler monitors structured perception signals in real time to switch between planners, and the MPC planner further optimizes trajectories over a receding horizon using 3D detections and HD maps. In other words, the framework keeps safety-critical trajectory planning/vehicle control within a conventional modular AD stack, and avoids low-level control being directly exposed to LLM hallucination risk.

\begin{figure*}[!t]
\centering
\includegraphics[width=2.0\columnwidth]{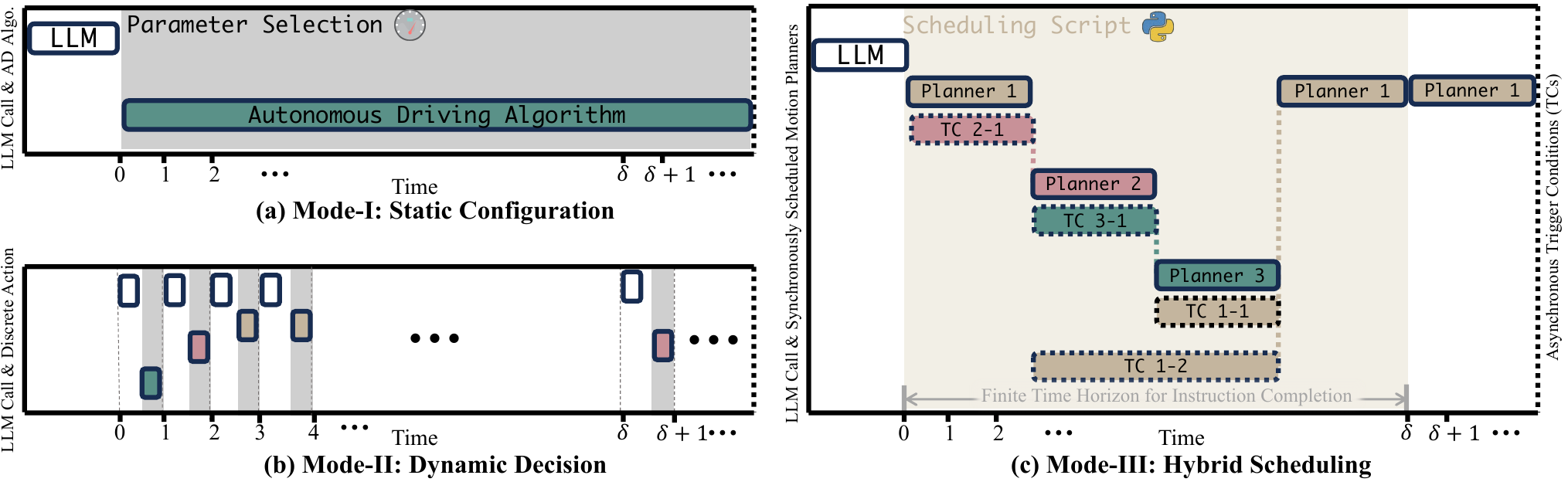} 
\caption{Categorization of LLM-Driven AD Methods from a Scheduling Perspective. (a) Mode I: LLM sets system parameters at startup, suitable for static preference configurations. (b) Mode II: LLM continuously engages in driving decisions in dynamic traffic. (c) Mode III (Ours): LLM produces a scheduling script in a single call, using asynchronous triggers for contextual adaptability. TC 2-1 denotes the \textit{Planner 2}'s \textit{1st} trigger condition.}
\label{fig:scheduling}
\end{figure*}

\subsection{Motion Planner Scheduling}
\label{sec:MethodSchedule}
Executing the behavior sequence $\left\{c_{i}\right\}$ demands seamless coordination between discrete decisions (e.g., \textit{when to switch from acceleration to a lane change}) and continuous control. Integrating both in a single policy $\pi_{\theta}$ is challenging yet vital for safe, efficient driving \cite{Hierarchical}. For this purpose, the framework employs a hierarchical policy: the LLM handles \emph{high-level discrete decisions}, while predefined motion planners manage \emph{low-level continuous control}.

\subsubsection{High-Level Discrete Decision-Making}

To elucidate the high-level decision-making of the framework, this study categorizes existing LLM-based AD methods into three modes (Figure \ref{fig:scheduling}), providing a structured taxonomy for formal comparison and discussion.\par

\emph{Mode I} uses the LLM to configure AD system parameters at startup, after which these parameters remain fixed \cite{ChatWithVehicle,Talk2Drive}. For example, given the instruction “\textit{Drive safely}”, an LLM may increase the safety term weight in the controller’s cost function. Nonetheless, due to the static configuration, this mode lacks the ability to make discrete, context-aware decisions in dynamic scenarios, making it unsuitable for maneuver-level instructions requiring sequential and conditional behavior switching.\par

\emph{Mode II} enables the LLM to make continuous decisions during driving. Such systems dynamically select discrete actions \cite{DiLu, DriveMLM}, tune AD parameters \cite{LanguageMPC}, or emit low-level control signals \cite{AutoVLA}, allowing robust management of evolving traffic conditions. This flexibility, however, increases computational overhead and latency due to frequent LLM queries. Our experiments also show that maintaining decision coherence throughout behavior sequence execution poses a new challenge for Mode II.\par

\emph{Mode III}, adopted by the proposed framework, executes the driving behavior sequence $\left\{c_{i}\right\}$ with a single LLM invocation while maintaining adaptability to evolving traffic. The LLM generates the executable script \textit{in a single pass}, which (i) schedules multiple motion planners to enact $\left\{c_{i}\right\}$ sequentially, and (ii) sets asynchronous triggers that monitor the scene graph and activate planner switches based on real-time conditions (e.g., \textit{when the gap exceeds 20 meters and..., switch from deceleration to a right lane change}). This hybrid mode achieves low overhead of Mode I and high contextual responsiveness of Mode II through script-based planner scheduling. \par

\subsubsection{Low-Level Continuous Control}
Following high-level LLM decisions, the cascaded motion planner and controller finally translate them into continuous control signals. The framework first invokes the behavior-specific MPC-based planner to generate trajectories, then applies a Linear Quadratic Regulator (LQR) for continuous control. At each step, MPC performs receding-horizon trajectory optimization with an explicit vehicle model, offering clear interpretability. The behavior-specific planner suite comprises: \textit{Lane Keeping}, \textit{Left Lane Change}, \textit{Right Lane Change}, \textit{Acceleration}, and \textit{Deceleration}.

This decoupled design brings the following advantages: (i) Expertise Domain Alignment: Constrain LLMs to high-level, discrete decisions while delegating low-level, continuous control to verifiable controllers. This keeps each component within its expertise domain, and prevents language-trained, probabilistic LLM from directly generating numerical, safety-critical control signals \cite{LM4TS_1,LM4TS_2}. (ii) Decision-Making Traceability: Human-readable scripts serve as an interface, enabling a transparent mapping from LLM textual reasoning to executed actions, simplifying inspection, debugging, and validation by developers or external auditors. (iii) Safety Robustness to Latency: Our decoupled design safeguards safety through a fast schedule–plan–control loop, thus reducing the impact of LLM inference latency—a critical factor for LLM-in-the-loop AD methods \cite{LLMAVSurvey}.  
\section{POINT Benchmark}

\begin{figure}[!t]
\centering
\includegraphics[width=1.0\columnwidth]{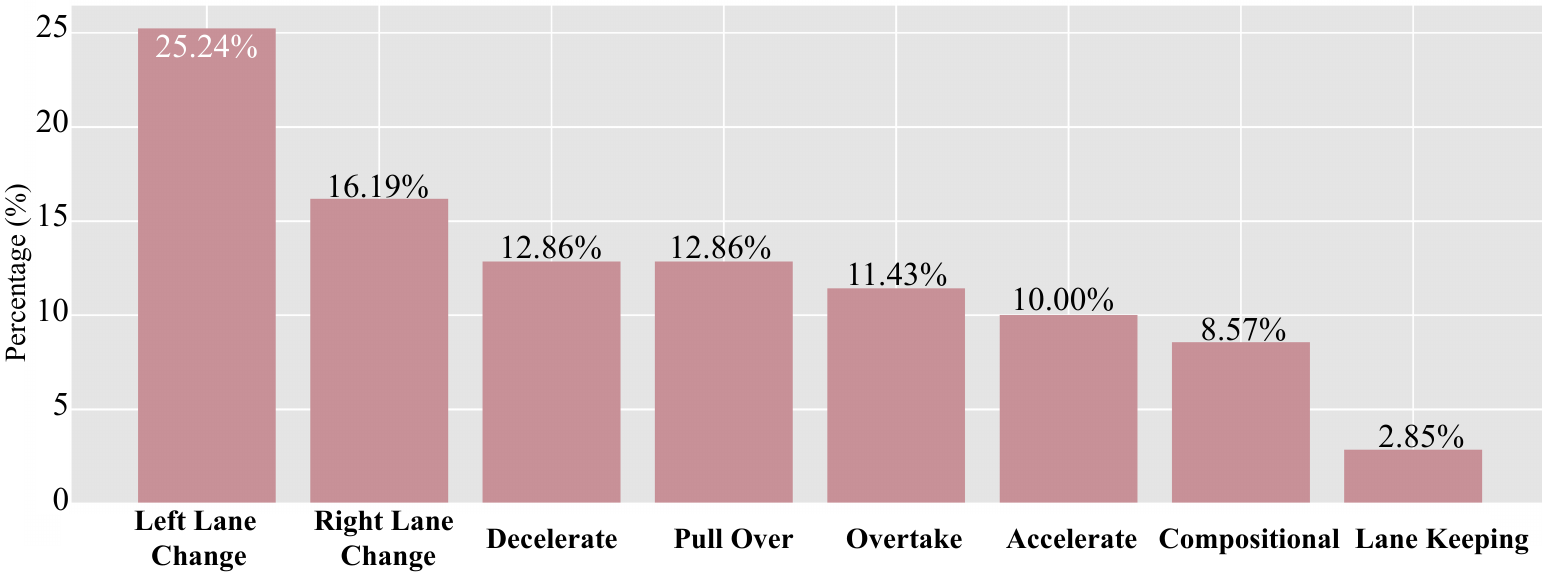} 
\caption{Instruction Intent Distribution.}
\label{fig:POINT}
\end{figure}

POINT comprises the nuPlan simulator, open-ended instructions, closed-loop evaluation metrics, and multiple competitive baselines.

\subsection{nuPlan Simulator}
POINT leverages the hybrid nuPlan simulator as its testing platform. nuPlan is the first publicly available simulator for real-world motion planning, designed to facilitate prototyping and evaluation of AD methods in urban settings \cite{Nuplan}. Built on 1,300 hours of real-world driving data, it reconstructs urban layouts, traffic patterns, and dynamic object states to provide high-fidelity simulation scenarios.\par

\subsection{Open-Ended Instructions}

POINT consists of 1,050 instructions paired with corresponding simulation‑initialization data. Initially, real-world instructions were collected, and commercial large-language models (such as ChatGPT and Gemini) were then used to generate additional instructions at scale. All the instruction-simulation pairs undergo rigorous manual screening for quality and relevance. To evaluate how well LLM understand open-ended instructions, the prompts used for instruction generation enforce conversational phrasing while suppressing explicit intent statements. Figure \ref{fig:POINT} reports high-level intent categories, where around 70\% of instructions involve risky lateral maneuvers (e.g., lane change, overtaking, and pulling over), supporting focused evaluation in high-stakes urban scenarios.\par

\subsection{Evaluation Metrics}
The benchmark evaluates short‑term instruction execution in urban traffic. Unlike standard AD tasks, it often requires instruction‑conditioned, high‑risk maneuvers (e.g., merging into heavy traffic). Accordingly, the evaluation focuses on task completion, safety, and rule compliance.\par

Specifically, \textit{task-related} metrics include: 1) Intent Recognition – fraction of instructions correctly parsed. 2) Instruction Realization – fraction of instructions successfully executed. \textit{Safety-related} metrics include: 3) Collision Avoidance – fraction of scenarios finished collision‑free. 4) TTC – minimum time‑to‑collision margin. \textit{Compliance-related} metrics include: 5) Drivable Area – time ratio within map‑defined drivable space. 6) Speed Limit Score – time ratio adhering to posted limits. 7) Direction Consistency – time ratio traveling in the correct lane direction. An \textit{efficiency-related} metric is also considered for specialized methods, i.e., 8) Expert Trajectory Progress – distance covered relative to a human expert.\par

\subsection{Baseline Methods}
The baseline methods of POINT include both specialized AD methods and instruction-realization methods. Specialized methods are tasked with following global paths derived from expert demonstration trajectories, whereas instruction-realization methods prioritize executing passenger instructions, often deviating from the global paths.

The specialized methods are as follows: 1) LogReplay \cite{Nuplan} replays the logged expert trajectories. 2) IDM \cite{IDM} is a longitudinal controller focusing on intra-lane car-following. 3) DiLu+, an extension of Mode-II DiLu \cite{DiLu} originally developed for discrete action selection in Highway-Env, is proposed in this study as a baseline method. It selects motion planners at 1 Hz, enabling continuous control in nuPlan. 4) PDM \cite{PDM} represents the SOTA solution for nuPlan closed-loop simulation.\par

The instruction-realization methods are as follows: 5) Diffusion-ES \cite{Diffusion-ES} combines LLM with black-box diffusion models in the Mode-III paradigm, where the LLM modifies the objective function and test-time optimization further directs trajectory generation. 6) DiLu++, another introduced Mode-II baseline, enhances DiLu+ by incorporating historical action and environment information into the LLM's input, thus enabling instruction realization.

\begin{table*}[!t]
\caption{Quantitative Comparison of Instruction Realization Performance. All LLM-based methods use a shared backbone, with average metrics computed across multiple random seeds. Each metric is in the [0, 1] range.}
\label{table:Qualitative}
\centering
\setlength{\tabcolsep}{0.3mm}
\begin{threeparttable}
\resizebox{2.0\columnwidth}{!}{
\begin{tabular}{ccccccccc}
\toprule[2pt]
\multicolumn{1}{c|}{\textbf{Method}} &
  \textbf{Categorization} &
  \textbf{Realization $\uparrow$} &
  \textbf{Collision $\uparrow$} &
  \textbf{TTC $\uparrow$} &
  \textbf{Drivable $\uparrow$} &
  \textbf{Speed $\uparrow$} &
  \textbf{Direction $\uparrow$} &
  \textbf{Progress $\uparrow$} \\ \hline
\multicolumn{9}{c}{\cellcolor[HTML]{ECF4FF}\textbf{Specialized AD Methods}}                                  \\
\multicolumn{1}{c|}{LogReplay}    & Expert Demonstration & -    & 0.86 & 0.84 &
\textbf{1.00} & 0.98 & \textbf{1.00} & \textbf{1.00} \\
\multicolumn{1}{c|}{IDM}          & Rule-Based        & -    & 0.87 & 0.76 & 0.90 & \textbf{1.00} & \underline{0.98} & 0.91 \\
\multicolumn{1}{c|}{PDMClosed}    & MPC-Based         & -    & 0.97 & \underline{0.86} & \underline{0.98} & \textbf{1.00} & \textbf{1.00} & \underline{0.92} \\
\multicolumn{1}{c|}{DiLu+}        & LLM+MPC           & -    & \textbf{1.00} & 0.84 & 0.96 & \underline{0.99} & \textbf{1.00} & \underline{0.92} \\ \hline
\multicolumn{9}{c}{\cellcolor[HTML]{ECF4FF}\textbf{Instruction-Realization Methods}}                         \\
\multicolumn{1}{c|}{Diffusion-ES} & LLM+Data-Driven   & 0.28 & 0.82 & 0.80 & 0.80 & \underline{0.99} & \textbf{1.00} & 0.77 \\
\multicolumn{1}{c|}{DiLu++}       & LLM+MPC           & \underline{0.51} & 0.92 & 0.73 & 0.96 & 0.97 & \textbf{1.00} & 0.87 \\
\rowcolor[HTML]{EFEFEF} 
\multicolumn{1}{c|}{\cellcolor[HTML]{EFEFEF}Ours} & LLM+MPC           & \textbf{0.84} & \underline{0.99} & \textbf{0.88} & 0.97 & \textbf{1.00} & \textbf{1.00} & 0.82 \\
\toprule[2pt]
\end{tabular}
}
\end{threeparttable}
\end{table*}

\begin{figure}[!t]
\centering
\includegraphics[width=1.0\columnwidth]{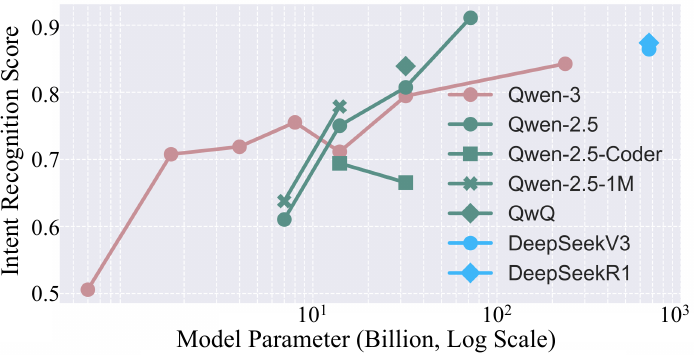} 
\caption{Intent Recognition Score of Various LLMs.}
\label{fig:IRScore}
\end{figure}

\section{Experiments}
To assess open-ended instruction realization and mitigate model bias \cite{ModelBias}, this study generates instructions using commercial LLMs, while the evaluation targets open-source LLM families such as Qwen \cite{Qwen} and DeepSeek \cite{DeepSeek}. Experiments are conducted on a workstation with Intel Xeon Gold 5220 CPUs and NVIDIA A40 GPUs. LLMs operated at their default conversational temperature. Baseline methods were run using the hyperparameters and checkpoints recommended in their original papers or project repositories to ensure fairness.

\subsection{Quantitative Evaluation}

\textbf{Intent Recognition}: Figure \ref{fig:IRScore} shows that intent recognition accuracy improves with LLM scale. Only the large-volume models—Qwen-2.5-72B, DeepSeekV3, and DeepSeek-R1—exceed 85\%, underscoring that interpreting open-ended instructions is a non-trivial task. Figure \ref{fig:IRScore} also indicates that reasoning mechanisms and larger context windows further improve performance.

\begin{figure*}[!t]
\centering
\includegraphics[width=2.0\columnwidth]{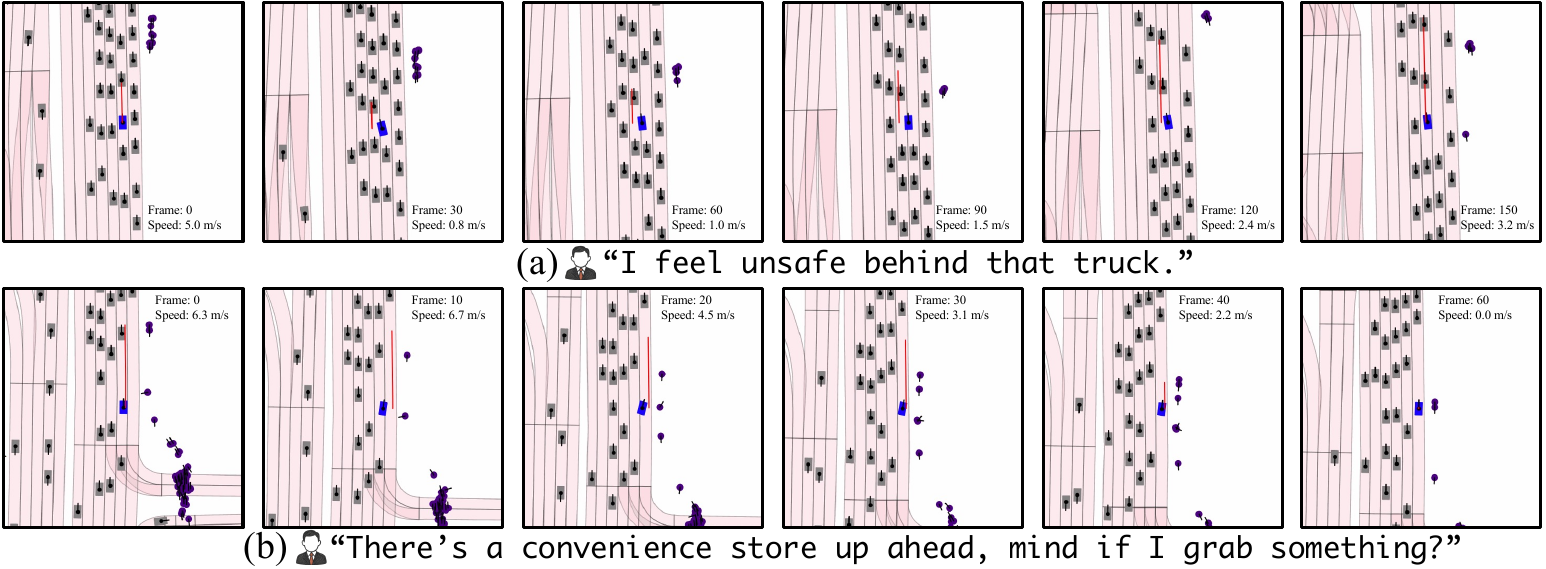} 
\caption{Key Frames in Instruction Execution Process. Animated results are available in the \textcolor{magenta}{\textit{Supplementary Material}}.}
\label{fig:case_study}
 \end{figure*}

\begin{figure}[!t]
\centering
\includegraphics[width=1.0\columnwidth]{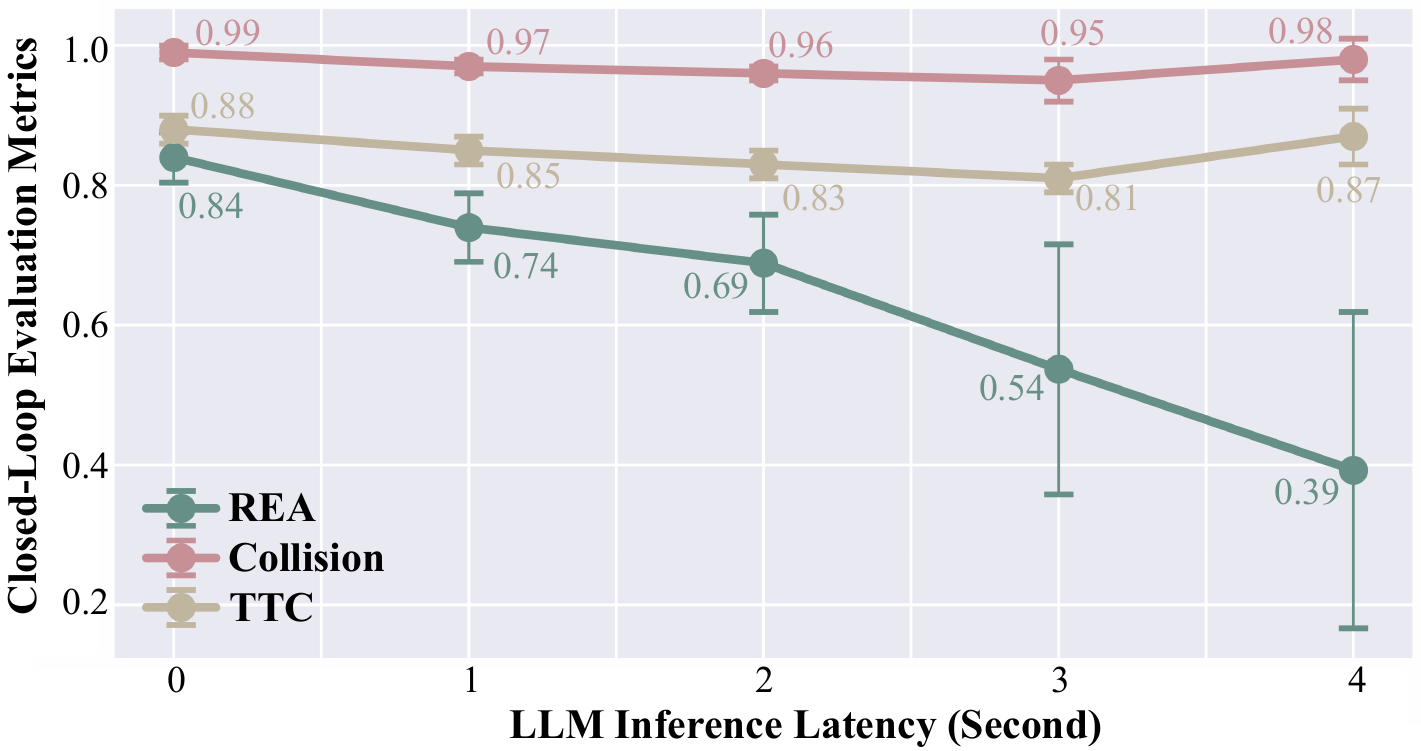} 
\caption{Latency Sensitivity Analysis.}
\label{fig:latency}
\end{figure}

\textbf{Instruction Realization}: Table \ref{table:Qualitative} compares the proposed framework with specialized and instruction-realization baselines. For fair comparison, all LLM-based baselines use the same DeepSeekV3 backbone and identical behavior sequences per intent-scenario pair.

Against specialized methods, the framework achieves leading safety and compliance. Expert Progress is lower due to occasional instruction-driven deviations from global paths. With the introduced motion planners, the proposed DiLu+ attains state-of-the-art results in Collision Avoidance and Expert Progress, showcasing the potential of combining high-level LLM with low-level planners for driving tasks.\par

Notably, it is non-trivial that our framework can execute risky instructions while matching the safety of specialized AD methods that do not follow instructions. This stems from our decoupled design, in which the fast MPC-based control loop ensures instruction realization never overrides vehicle safety.

Among instruction-realization methods, the framework achieves the highest Instruction Realization score of 0.84, balancing task execution and safety. DiLu++ ranks second with 0.51 but occasionally ignores past actions and context, \textit{leading to incoherent decisions} such as redundant lane changes. Despite being trained on extensive expert trajectories, Diffusion-ES underperforms in both effectiveness and safety, sometimes producing near-stationary trajectories.

\begin{table}[!t]
\caption{Ablation Study on Traffic Context and LLM-Guided Planner (PL) Scheduling.}
\label{table:AblationStudy}
\centering
\setlength{\tabcolsep}{0.3mm}
\begin{threeparttable}
\resizebox{1.0\columnwidth}{!}{
\begin{tabular}{cccc}
\toprule[2pt]
\multicolumn{1}{c|}{\textbf{Methods}} & \textbf{REC/REA $\uparrow$} & \textbf{Collision $\uparrow$} & \textbf{TTC $\uparrow$} \\ \hline
\rowcolor[HTML]{ECF4FF} 
\multicolumn{4}{c}{\cellcolor[HTML]{ECF4FF}{\color[HTML]{000000} \textbf{Intent Recognition (REC)}}} \\
\multicolumn{1}{c|}{Ours w.o. Context} & \underline{0.78}  & - & - \\
\rowcolor[HTML]{EFEFEF} 
\multicolumn{1}{c|}{\cellcolor[HTML]{EFEFEF}Ours} & \textbf{0.86}  & - & - \\ \hline
\rowcolor[HTML]{ECF4FF} 
\multicolumn{4}{c}{\cellcolor[HTML]{ECF4FF}\textbf{Instruction Realization (REA)}} \\
\multicolumn{1}{c|}{Lane Keeping PL} & 0.17 & \underline{0.97}  & 0.86  \\
\multicolumn{1}{c|}{Left Lane Change PL} & \underline{0.18}  & 0.95  & 0.75  \\
\multicolumn{1}{c|}{Right Lane Change PL} & 0.14  & 0.98  & \underline{0.90} \\
\multicolumn{1}{c|}{Acceleration PL} & 0.13 & 0.57 & 0.38 \\
\multicolumn{1}{c|}{Deceleration PL} & 0.12 & \textbf{0.99} & \textbf{0.97}\\
\rowcolor[HTML]{EFEFEF} 
\multicolumn{1}{c|}{\cellcolor[HTML]{EFEFEF}PL Scheduling (Ours)} & \textbf{0.84}& \textbf{0.99} & 0.88  \\ 
\toprule[2pt]
\end{tabular}
}
\end{threeparttable}
\end{table}

\textbf{Ablation Study}: Table \ref{table:AblationStudy} shows performance drops in intent recognition and instruction realization when ablating traffic context and planner scheduling. Including contextual cues increases performance by about 10\% for DeepSeek-V3, providing a more accurate instruction interpretation. The high‑level LLM scheduling effectively coordinates the low‑level motion planners, increasing task‑completion rates without compromising safety.\par

\textbf{Latency Sensitivity Analysis}: Considering LLMs' non-negligible inference overhead, this experiment introduces controlled delays into the LLM decision process and track the instruction-realization score (REA) and safety indicators. Figure \ref{fig:latency} shows that increasing latency causes a gradual decline in REA while safety metrics remain stable.\par

This robustness also stems from the decoupled framework design: LLM is queried at low frequency to produce a global scheduling scheme, whereas a real-time, feedback-driven inner loop continuously runs at high frequency and enforces emergency behaviors. This separation yields considerable tolerance for LLM inference delays.

\subsection{Qualitative Evaluation}
Figure \ref{fig:case_study} illustrates key moments from the framework's instruction execution. In (a), the system infers an implicit lane-change intent, decelerates, and merges into a slower, denser lane. In (b), it responds to a pull-over request by changing lanes and stopping safely.

\section{Conclusion \& Discussion}
This study presents a LLM-enabled, scheduling-centric framework to execute open-ended instructions, decoupling instruction interpretation, planner scheduling, and motion planning across different timescales while ensuring a transparent decision-making chain from high-level decision to low-level control. Due to the lack of testbeds, it also introduces POINT, a high-fidelity benchmark with multiple closed-loop metrics and diverse baselines. Experiments show that: (i) Interpreting open-ended instructions is non-trivial and requires highly capable, large-scale LLMs; adding explicit reasoning and traffic context improves instruction understanding. (ii) With a single LLM query, the proposed framework achieves an instruction realization score of 0.84, outperforms the baselines by 64\% to 200\%, meets the safety standards of specialized AD methods, and remains safety-robust to LLM inference delays.\par

In LLM-involved instruction realization, passenger instructions must be handled cautiously because LLM outputs are probabilistic and can hallucinate, and passengers may have limited system understanding or driving experience. The framework therefore restricts instruction-conditioned, LLM-generated scripts to schedule-stage transitions, while all vehicle control is managed by atomic planners. This creates a passenger/LLM-in-the-loop safety mechanism that ensures every executed action comes from safety-constrained atomic planners despite risky instructions, hallucinations, or LLM latency.\par

Although the framework relies on a predefined library of atomic planners, it remains flexible through composition. The supported instruction task space scales as ``\textit{triggers} $\times$ \textit{planners} $\times$ \textit{temporal ordering}", so a small planner set can cover diverse instructions. Meanwhile, adding verified planners and triggers can expand the task space combinatorially, enabling rapid capability growth with low marginal integration cost.\par

Despite these gains, future efforts are still needed before practical deployment: (i) \textit{Visual Integration}: Since visual inputs convey richer semantics than text \cite{EyeDirection}, integrating VLM for instruction understanding remains critical. (ii) \textit{Simulation Augmentation}: nuPlan lacks ego-view image rendering, constraining closed-loop evaluation of VLA methods. Integrating advances such as 3D Gaussian splatting \cite{OmniRe} for urban scenario reconstruction is essential for a comprehensive assessment. (iii) \textit{Trigger Expressiveness}: Scheduling responsiveness is bounded by the expressiveness of asynchronous triggers. Trial-and-error learning and adaptive re-invocation remain key to improving generality.\par

\section{Acknowledgments}
This work was supported by National Natural Science Foundation of China under Grant (62573209), Development and Reform Commission Foundation of Jilin Province under Grant (2024C003), Doctoral Student Research Innovation Capacity Enhancement Program of the Education Department of Jilin Province under Grant (JJKH20250236BS), and the Agency for Science, Technology and Research (A*STAR) under its MTC Programmatic Funds (Grant No. M23L7b0021).

{
    \small
    \bibliographystyle{ieeenat_fullname}
    \bibliography{main}
}

\end{document}